\documentclass[runningheads]{llncs}
\usepackage{graphicx}
\usepackage{amssymb}
\usepackage{array}
\usepackage[misc,geometry]{ifsym}
\usepackage{color}
\begin{document}
\title{Impressions2Font: Generating Fonts by Specifying Impressions}
\titlerunning{Generating Fonts by Specifying Impressions}

\author{Seiya Matsuda\inst{1}(\Letter) \and
Akisato Kimura\inst{2} \and
Seiichi Uchida\inst{1}\orcidID{0000-0001-8592-7566}}
\authorrunning{S. Matsuda et al.}
\institute{Kyushu University, Fukuoka, Japan
\email{seiya.matsuda@human.ait.kyushu-u.ac.jp}\and
NTT Communication Science Laboratories, NTT Corporation, Japan}

\maketitle              
\begin{abstract}
Various fonts give us various impressions, which are often represented by words.
This paper proposes Impressions2Font (Imp2Font) that generates font images with specific impressions. Imp2Font is an extended version of conditional generative adversarial networks (GANs). More precisely, Imp2Font accepts an arbitrary number of impression words as the condition to generate the font images. These impression words are converted into a soft-constraint vector by an impression embedding module built on a word embedding technique. Qualitative and quantitative evaluations prove that Imp2Font generates font images with higher quality than comparative methods by providing multiple impression words or even unlearned words.
\end{abstract}
%
%
\section{Introduction} \label{sec:intro}
\begin{figure}[t]
    \centering
    \includegraphics[width=0.8\linewidth]{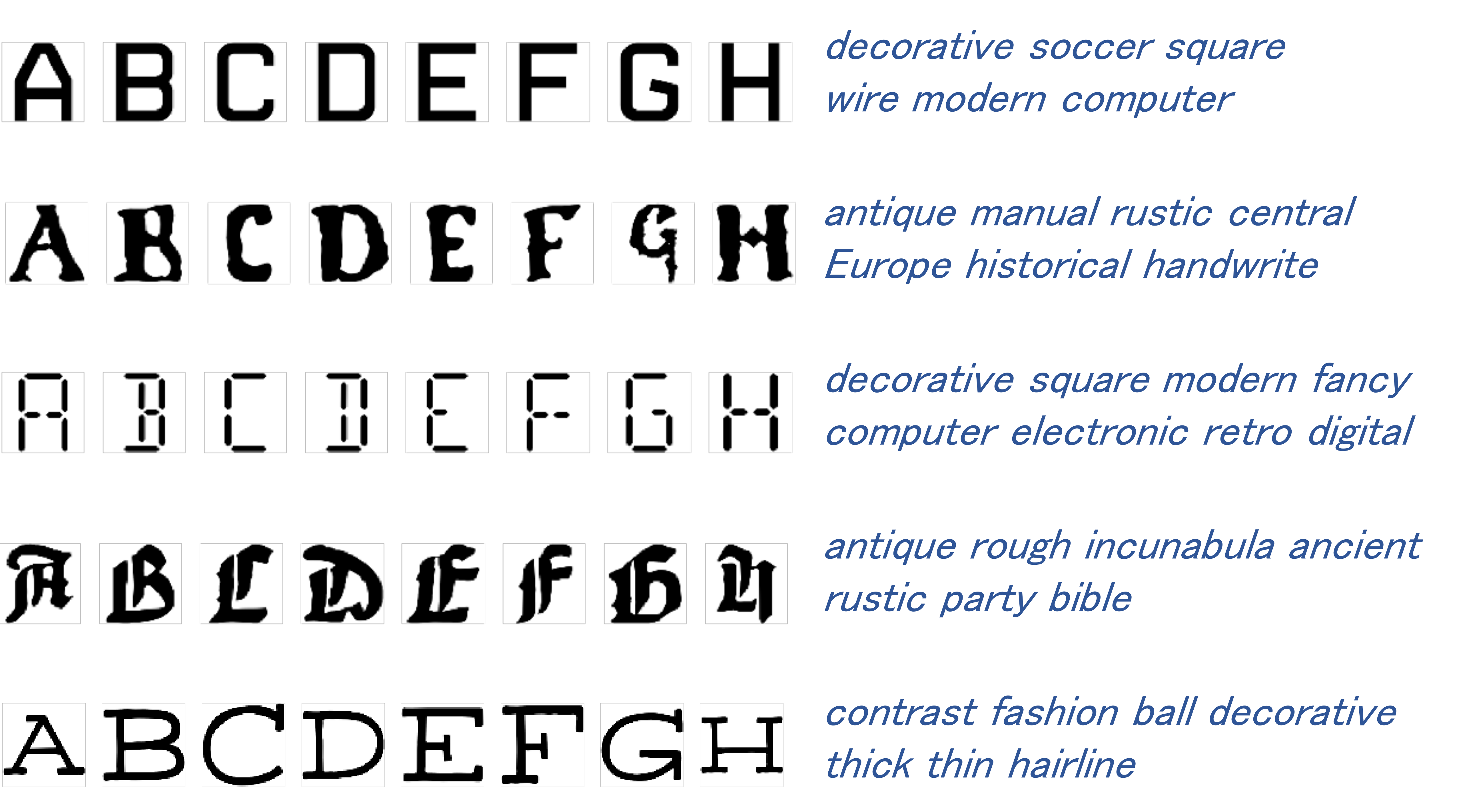}\\[-3mm]
    \caption{Fonts and their impression words from MyFonts dataset~\cite{Chen2019large}.}
    \label{fig:shape-and-impression}
\end{figure}

One of the reasons we have huge font design variations is that each design gives a different impression. For example, fonts with serif often give a {\it formal} and {\it elegant} impression. Fonts without serif (called sans-serif) often give a {\it frank} impression. Decorative fonts show very different impressions according to their decoration style. Fig.~\ref{fig:shape-and-impression} shows several fonts and their impression words. These examples are extracted from MyFonts dataset~\cite{Chen2019large}, which contains 18,815 fonts and 1,824 impression words (or tags) from {\tt MyFonts.com}.  \par
%
The ultimate goal of our research is to provide font images well suited for any given impression.
One possible solution is tag-based font image retrieval; we can retrieve any existing font images by impression queries if they are appropriately annotated with impression tags. However, such annotations will require a substantial amount of effort. Even with full impression annotations, we can obtain neither any font images with new text tags nor completely new font images.\par
Based on the above discussion, one of the best ways to the goal is to generate font images with specific impressions
\footnote{In this paper, we use the term ``impression'' in a broader meaning; some impression is described by words that relate more to font shapes, such as sans-serif, rather than subjective impression.}. For example, we aim to generate a font image with an {\it elegant} impression or a font with a {\it scary} impression. Such a font image generator will be very useful to various typographic applications. In addition, it is also meaningful to understand the relationship between font style and impression.\par
\begin{figure}[t]
    \centering
    \includegraphics[width=1.0\linewidth]{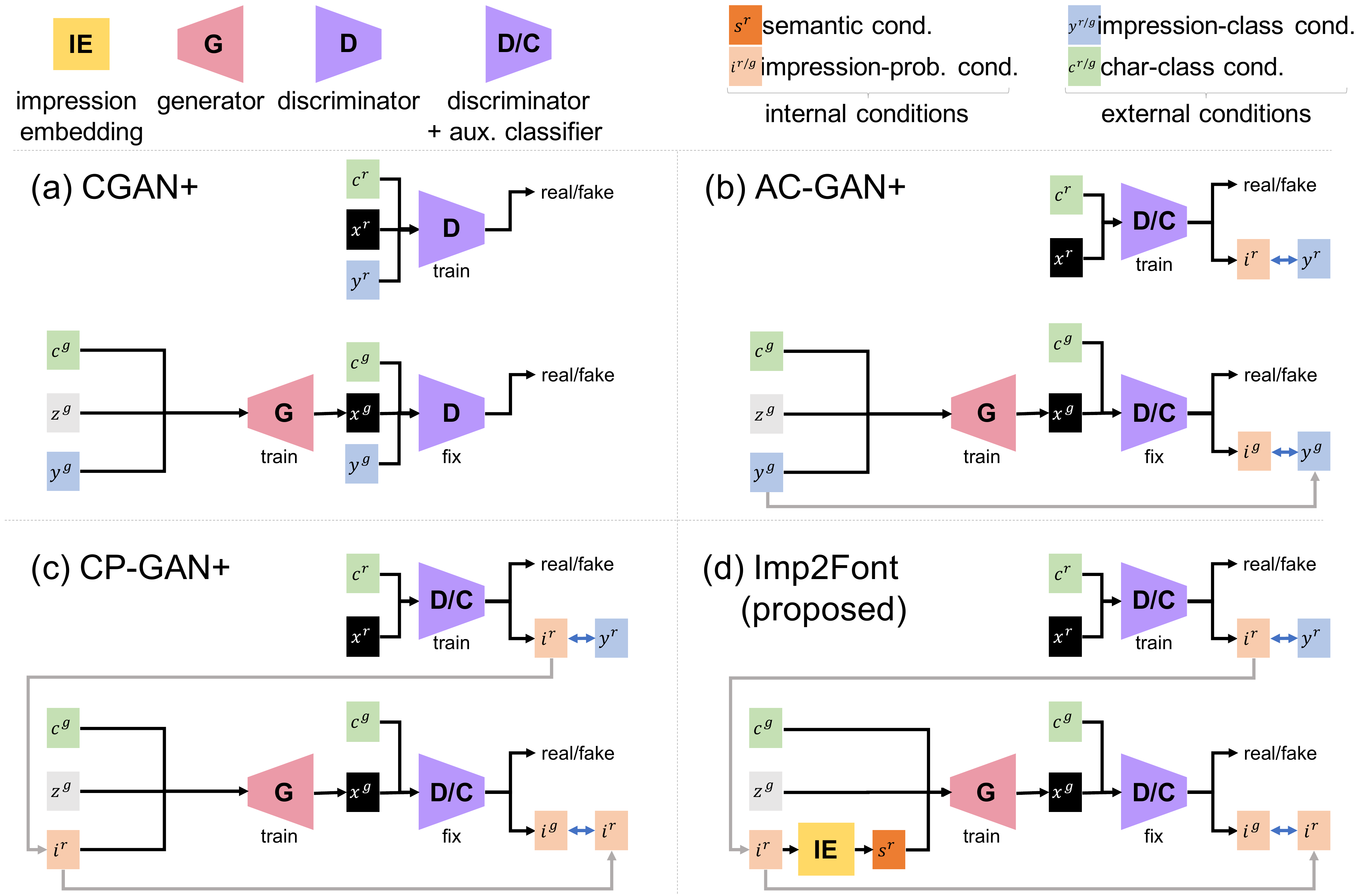}\\[-3mm]
    \caption{(a)-(c) three representative conditional GANs, CGAN~\cite{mirza2014conditional}, AC-GAN~\cite{odena2017conditional}, and CP-GAN~\cite{kaneko2019CP-GAN}, and (d) the proposed Impressions2Font (Imp2Font, for short). For (a)-(c), we add `+' to their name, like CGAN+, to indicate that they are an extended version from their original model for dealing with the character class condition, $c^{r/g}$.}
    \label{fig:proposed_model}
\end{figure}
Fig.~\ref{fig:proposed_model}~(a) shows a naive GAN model for generating a font image with a specific impression. It is based on Conditional Generative Adversarial Networks (CGAN)~\cite{mirza2014conditional}. Unlike the traditional GAN~\cite{goodfellow2014generative}, CGAN can control the output images by using an external condition  $y^{r/g}$, where the superscripts $r$ and $g$ suggest that the condition is prepared for the real data and the generated data, respectively.
For our purpose, we accept a one-hot representation with $K$ impression word vocabulary as an external condition $y^{r/g}$ and try to generate a font image with an impression specified by the one-hot representation.
Here, we note that the CGAN in Fig.~\ref{fig:proposed_model}~(a) is slightly extended from its original model to incorporate another condition $c^{r/g}$ to specify the character class (`A'-`Z').  We thus call this extended CGAN as CGAN+ hereafter, for clarifying that it is not an original CGAN.\par
This naive realization font image generation by CGAN+ still has a large room for improvement. The most significant problem is that it treats different impression words just as different conditions. For example, a pair of synonyms, such as {\it large} and {\it big}, must work as almost the same condition for generating font images. However, they are treated as two totally different one-hot vectors in CGAN+, like the conditions of totally different words, {\it heavy} and {\it elegant}.\par
In this paper, we propose a novel network model, called Impressions2\allowbreak Font (Imp2Font, for short), for generating font images by specifying impression words. Fig.~\ref{fig:proposed_model}~(d) shows the overall structure of Imp2Font, whose components are detailed in the later section. Although its structure is inspired by AC-GAN~\cite{odena2017conditional} (Fig.~\ref{fig:proposed_model}~(b)) and CP-GAN~\cite{kaneko2019CP-GAN} (Fig.~\ref{fig:proposed_model}~(c)), Imp2Font has a novel module, called {\em impression embedding module}, to deal with various impressions more efficiently and more flexibly. \par
The impression embedding module enables us to deal with the mutual relationships among impression words by using their semantic similarity.
More specifically, this new module uses a word embedding technique to convert each word to its real-valued semantic vector, and it is trained with external large-scale text corpora.
It is well-known that word-embedding modules trained with large-scale text corpora will give similar semantic vectors to words with similar meanings.
Consequently, using this semantic vector representation as real-valued conditions, mutual relationships among multiple impression words will be naturally incorporated into the CGAN framework. \par
Based on the above discussion, the advantages of the impression embedding module for our font generation task are summarized as follows.
\begin{enumerate}
    \item We can specify {\em multiple} impression words for generating font images.
    \item We can generate font images for {\em unlearned} impression words that are not included in the vocabulary used in the training phase.
    \item The impression embedding module relaxes the data imbalance problem. Several impression words are rarely attached to fonts, and therefore difficult to associate them with font images. However, with the impression embedding module, similar and major impression words will help training for the rare impression words by sharing examples automatically. 
    \item The impression embedding module relaxes the noisy-label problem. Impression words attached to fonts are often very subjective and noisy. For example, two rather inconsistent impressions, ``modern'' and ``retro'', are attached to the third font of Fig.~\ref{fig:shape-and-impression}. Since the proposed module converts the hard one-hot constraint (showing the labeled impression) into a soft real-valued constraint internally, it can weaken the effect of noisy labels. 
\end{enumerate}\par

In the experiment, we train Imp2Font by font images and their impressions from 
the dataset (hereafter called MyFonts dataset) published by Chen et al.\cite{Chen2019large} 
 and then 
generate font images with various impressions. We also conduct qualitative and quantitative evaluations and comparisons with other CGANs. We also generate font images by specifying multiple impression words and unlearned impression words.\par
The main contributions of this paper are summarized as follows:
\begin{itemize}
    \item This paper proposed a novel GAN-based font generation method, called Impressions2Font, which can generate font images by specifying the expected impression of the font by possibly multiple words. 
    \item Different from \cite{wang2020attribute2font} where 37 impression words are pre-specified, this is the first attempt to cope with arbitrary impression words, even unlearned words.
    \item The experimental results and their quantitative and qualitative evaluations show the flexibility of the proposed method, as well as its robustness to the data imbalance problem and the noisy-label problem.
\end{itemize}

\section{Related Work}\label{sec:related}
\subsection{Font style and impression}
Research to reveal the relationship between font style and its impression has an about 100-year history, from the pioneering work by Poffenebrger~\cite{poffenberger1923study}. 
As indicated by the fact that this work and its succeeding work~\cite{davis1933determinants} were published in {\it American Psychological Association}, such research was mainly in the scope of psychology for a long period. From around the 1980s, fonts have also been analyzed by computer science research. After tackling a simple bold/italic detection task (for OCRs), font identification, such as Zramdini et al.~\cite{zramdini1998optical}, becomes one of the important tasks. DeepFont~\cite{wang2015deepfont} is the first deep neural network-based attempt for the font identification task. Shirani et al.\cite{Shirani2020} analyze the relationship between font styles and linguistic contents.
\par
O'Donovan et al.~\cite{o2014exploratory} is one of the pioneering works on the relationship between font style and its impression in computer science research.
They used a crowd-sourcing service to collect impression data for 200 fonts. The degrees of 37 attributes (i.e., impressions, such as dramatic and legible) are attached to each font by the crowd-workers. They have provided several experimental results, such as the attribute estimation by a gradient boosted regression.\par
Recently, Chen et al.~\cite{Chen2019large} publish the MyFonts dataset by using the data from {\tt Myfonts.com} and use it for a font retrieval system. They also use a crowd-sourcing service to clean up this noisy dataset as much as possible. It contains 18,815 fonts annotated with 1,824 text tags (i.e., the vocabulary size of impression words). We will use this dataset for training our Imp2Font; this large vocabulary helps us to generate font images by specifying arbitrary impression words.\par
%
\subsection{GAN-based font generation}
\par
Recent developments in generative adversarial networks (GANs) have contributed significantly to font generation. Lyu et al.~\cite{lyu2017auto} propose a GAN-based style transfer method for generating Chinese letters with various calligraphic styles. Hayashi et al.\cite{hayashi2019glyphgan} have proposed GlyphGAN, which is capable of generating a variety of fonts by controlling the characters and styles. As a related task, font completion has also been investigated~\cite{jiang2017dcfont,azadi2018multi,Zhu2020,Cha2020}, where methods try to generate letter images of the whole alphabet in a font by using a limited number of letter images of the font (i.e., so-called few-shot composition). All of the previous methods for font completion employ GAN-based models as well.\par
Recently, Wang et al.\cite{wang2020attribute2font} proposed a GAN-based font generation model that can specify font impressions.
In this work, 37 impressions used in the dataset by O'Donovan et al.~\cite{o2014exploratory} are prepared to control the font shape.
Although their purpose is similar to our trial, there exist several differences as follows:
First, their impressions are limited to 37, but ours are infinite in principle, thanks to our proposed impression embedding module.
Second, their method needs to specify the strength values of all the 37 impressions. In contrast, ours does not need~\footnote{These two differences make it very difficult to fairly compare Wang et al.\cite{wang2020attribute2font} and our proposed method. }.
Third, their method does not consider the semantics of each impression word, whereas ours does --- this leads to our main contributions, as listed in Section~\ref{sec:intro}.

\subsection{Conditional GANs}
Conditional GANs have been proposed to generate images that fit a specific condition, such as class labels. 
The original Conditional GAN (hereafter, called CGAN) was proposed by Mizra et al.~\cite{mirza2014conditional} for controlling the generated image by giving a class label to the generator and the discriminator. Fig.~\ref{fig:proposed_model}~(a) is CGAN with a slight modification that can deal with the character class condition $c^{r/g}$, in addition to the main condition $y^{r/g}$ related to the semantics of impression words. In this paper, a plus symbol `+' is attached to the GAN's name (e.g., CGAN+, AC-GAN+, CP-GAN+) to clarify that the character class condition augments them. 
\par
Odena et al.~\cite{odena2017conditional} proposed Auxiliary Classifier GAN
(AC-GAN), where output images are classified by an auxiliary classifier sharing the feature extraction layers with the discriminator. This auxiliary classifier makes it possible to evaluate how a generated image fits a given condition $y^{r/g}$.  Fig.~\ref{fig:proposed_model}~(b) shows AC-GAN plus character class conditions. In our font generation scenario, the auxiliary classifier will output a $K$-dimensional vector $i^{r/g}$ whose element corresponds to the likelihood of an impression, and it will be trained so that the one-hot impression condition $y^{r/g}$ is as close to $i^{r/g}$ as possible.\par
Kaneko et al.~\cite{kaneko2019CP-GAN} proposed Classifier's Posterior GAN (CP-GAN), which is an improved version of AC-GAN.
CP-GAN also has an auxiliary classifier; however, the usage of its outputs is unique. As shown in Fig.~\ref{fig:proposed_model}~(c), an output $i^r$ of the auxiliary classifier is also used as an {\it internal} condition for a generator in CP-GAN+ along with {\it external} condition $c^g$. Since $i^r$ is a real-valued vector representing the likelihood of $K$ impression words, it works as a softer constraint than the one-hot condition $y^g$. It thus can realize a more flexible control of the generator during its training and testing. In fact, the vector $i^r$ can be seen as a multi-hot vector representing the strength of all $K$ conditions, and therefore we can generate images satisfying multiple conditions with certain strengths.\par
Although we developed our Imp2Font with inspiration from CP-GAN, the introduction of the impression embedding module realizes several essential differences between CP-GAN and ours. This new module works as an appropriate condition mechanism for our font generation task; especially, it can consider the mutual relationship among conditions and thus help to generate the fonts with arbitrary conditions even by the training data with imbalance and noisy labels. We will compare our method with CP-GAN+, AC-GAN+, and CGAN+, experimentally in the later section.

\begin{figure}[t]
    \centering
    \includegraphics[width=1.0\linewidth]{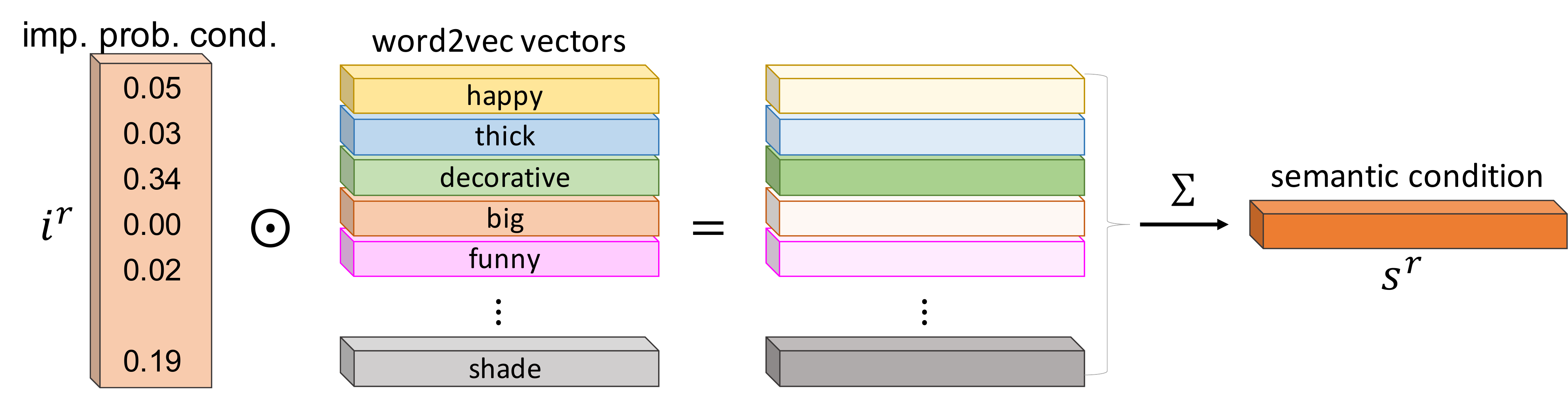}\\[-3mm]
    \caption{The architecture of Impression Embedding Module.}
    \label{fig:impression-embedding}
\end{figure}
\par

\section{MyFonts Dataset\label{sec:dataset}}
As noted above, we use the MyFonts dataset published by Chen et al.~\cite{Chen2019large}, which is comprised of 18,815 fonts. 
Since the dataset contains the dingbat fonts and the circled fonts, 
we removed them by manual inspection; as the result, we use
26 letter 
glyph images of `A' to `Z' from  17,202 fonts for training Imp2Font.
\par
The vocabulary size of the impression words (see Fig.~\ref{fig:shape-and-impression} for their examples) is 1,824
\footnote{As noted later, each impression word is converted to a semantic vector by word2vec~\cite{mikolov2013distributed}. Therefore, we remove too rare impression words that are not included even in the 3-million English vocabulary for training word2vec. This results in $K=1,574$ impression words that we used in the following. Note that an impression word with hyphenation is split into sub-words, and then its semantic vector is derived by taking the sum of the semantic vectors of the sub-words.}
and at most $184$ impression words are attached to each font. The examples in Fig.~\ref{fig:shape-and-impression} are collected from the MyFonts dataset.
\par
The MyFonts dataset has several problems for training a GAN-based font image generator. First, it has a data imbalance problem. For example, the impression word {\it decorative} is attached to 6,387 fonts, whereas 16 impression words (including {\it web-design}, {\it harmony}, etc.) is associated with only 10 fonts. Second, it contains noisy labels.
Since the impression words are attached by many people (including non-experts) and the impression of a font is often person-dependent, the impression words of a font are not very consistent. For example, two contradicting impressions, {\it thin} and {\it thick}, are attached to the bottom font of Fig.~\ref{fig:shape-and-impression}. Another interesting example is {\it modern} and {\it retro} of the second font; it might seem modern for an older person 
but retro for a young person. Third, the number of impression words, $K$, is still limited to 1,574, although it is far larger than 37 of \cite{o2014exploratory}. Since it is known that English word vocabulary 
for common use 
is more than 10,000, we might encounter out-of-vocabulary words in several applications, such as the font generation with specifying its impression by a ``sentence'' or a ``text.''
Fortunately, as noted in the later sections, our Imp2Font is robust to these problems by its word embedding module. 

\section{Impressions2Font --- Conditional GAN with Impression Embedding Module}
\subsection{Overview}
This section details our proposed Imp2Font, which can generate font images by specifying an arbitrary number of arbitrary impression words. As already seen in Fig.~\ref{fig:proposed_model}(d), Imp2Font accepts two external conditions, 
namely a character class $c^{r/g}$ and impression words $y^{r/g}$. 
The former is a 26-dimensional one-hot vector (for generating 26 English capitals, `A'-`Z'), and the latter is a $K$-dimensional one-hot vector, where randomly selected an impression word from the impression words annotated in the font at each epoch. We do not use the $K$-hot representation to satisfy $\sum{i^r}=1$ in the impression embedding module. 
The auxiliary classifier is trained so that its real-valued output $i^r$ becomes more similar to a one-hot ground-truth $y^r$, just like AC-GAN.\par
Inspired by CP-GAN, we use this $i^r$ also as an {\em internal condition} for training the generator. We call $i^r$ an impression probability condition. We then optimize the generator so that this impression probability condition $i^g$ of generated data corresponds with that $i^r$ of real data. This formulation allows our model to capture between-impression relationships in a data-driven manner and generate a font image conditioned on the impression specificity. \par
The impression probability condition $i^r$ relates to the highlight of the proposed method, i.e., the impression embedding module. More specifically, the condition $i^r$
is converted to another internal condition, called {\em semantic condition} $s^r$, 
by the module, as detailed in the next section~\ref{sec:impression-embedding}.
The semantic condition is a $D$-dimensional real-valued vector and specifies the impression to be generated. During the training phase of the generator, the semantic condition is fed to the generator. \par
%
\subsection{Impression embedding module\label{sec:impression-embedding}}
Fig.~\ref{fig:impression-embedding} shows the architecture of our impression embedding module. This module gives the $D$-dimensional semantic condition $s^r$, which is expected to be an integrated vector of the impressions specified by the impression probability condition, $i^r$. In the example of this figure, the probabilities of the impressions, {\it decorative} and {\it happy} are high (0.34) and low (0.05), respectively. Therefore, the semantic condition $s^r$ should represent more {\it decorative} and {\it happy} as the integrated impression. \par
In the impression embedding module, the semantic condition is derived as
\begin{equation}
 s^r = \sum_{k=1}^K i^r_k w_k,\label{eq:impression-embedding}
\end{equation}
where $i^r_k$ is the $k$-th element of $i^k$ and $w_k$ is a semantic vector 
of the $k$-th impression word. The semantic vector $w_k$ is expected to represent the meaning of the $k$-th impression word as a real-valued $D$-dimensional vector. Also, the $k$-th and $k'$-th words have similar semantic meanings, $w_k$ and $w_{k'}$ are also expected to be similar.\par
Although an arbitrary word embedding method can be employed for extracting semantic vectors, we introduce  word2vec~\cite{mikolov2013distributed}, which is a traditional but still useful method. We use a word2vec model pre-trained by the large-scale Google News dataset (about 100 billion words). The model outputs a $D=300$-dimensional semantic vector for each word in the 3-million English vocabulary.\par
Our formulation Eq.~(\ref{eq:impression-embedding}) of the impression embedding module offers several advantages for generating font images from impression words.\par
(1) Robustness against data imbalance: Let us assume that the impression word {\it fat} is rarely attached to font images, whereas a similar impression word {\it bold} is frequent. If a typical $K$-dimensional one-hot vector represents the semantic condition, it is hard to generate font images with {\it fat} impression due to the lack of its training samples. However, our semantic condition by Eq.~(\ref{eq:impression-embedding}) allows to transfer knowledge from font styles the {\it bold} impression to those with the {\it fat} impression, because we can expect $w_{\rm bold}\sim w_{\rm fat}$.\par
(2) Robustness to noisy labels:
Let us assume that three impression words, {\it bold}, {\it heavy}, and {\it thin}, are attached to a font. The impression {\it thin} seems to be inconsistent with the others, and thus it is expected to be a noisy label. Note that, as shown in Fig.~\ref{fig:shape-and-impression}, the MyFonts dataset contains such inconsistent annotations.
The impression class condition $y^r$ will become a one-hot vector which represents any of the three impression words;
however, the resulting semantic condition $s^r$ is less affected by the noisy-label {\it thin} than the other two words, because the meanings of {\it bold} and {\it heavy} are similar, that is, $w_{\rm bold}\sim w_{\rm heavy}$, and have a double impact in $s^r$.
\par
\subsection{Generating font images from the trained generator}
After the adversarial training of the generator and the discriminator (plus the auxiliary classifier), we can generate character images with specific impressions by giving the character-class condition $c^g$, the noise vector $z^g$, and an arbitrary semantic condition $s^g$. By changing $z^g$, we have various fonts even with the same impression $s^g$ and the same character class $c^g$.
\par
The condition $s^g$ is fed to the generator like $s^r$; however, different from the internal condition $s^r$, the condition $s^g$ is an external condition and specified by directly using arbitrary semantic vectors given by word2vec.  For example, if we want to generate a font with the impression {\it elegant}, we can use $w_{\rm elegant}$ as $s^g$. \par
The flexibility of setting the condition $s^g$ realizes
Imp2Font's strength that we can specify an arbitrary number of arbitrary impression words. For example, a font with multiple impressions is generated by setting $s^g$ as the sum of the corresponding semantic vectors. Moreover, it is possible to use the semantic vector of {\em unlearned} impression words as $s^g$; even if the impression word {\it gigantic} is unlearned, Imp2Font will generate a gigantic font with the help of some similar and learned words, such as {\it big}.
\par

\subsection{Implementation details}\label{sec:implementation}
Table~\ref{table:SpeedOfLight} shows the network structure of the generator and the discriminator in the proposed method. 
The structure of the generator and discriminator is based on commonly known DCGANs.
To avoid mode-collapse, the mode seeking regularization term by Mao et al.~\cite{mao2019mode} is employed for the generator. The Kullback-Leibler (KL) divergence is employed as the loss function of the auxiliary classifier. During the 100-epoch training, ADAM optimization was used with the learning rate at $0.0002$. 
The batch size is 512, and the generator and discriminator are updated alternately in a 1:5 ratio.\par
In the experiments, CGAN+, and AC-GAN+, and CP-GAN+ are used. For a fair comparison, we used the same setup for them, including the model seeking regularization. Since Imp2Font is a super-set of those methods (as indicated by Fig.~\ref{fig:proposed_model}), the following experiments can be seen as ablation studies. \par


\begin{table}[t]
 \caption{Network structure.\label{table:SpeedOfLight}}
 \begin{scriptsize}
 \centering
 \begin{minipage}[t]{0.45\textwidth}
 \vspace{0pt}
  \begin{tabular}{c}\hline
     {\bf Generator} \\ \hline
     $z^g \in \mathbb{R}^{300}\mathtt{\sim}N(0, I)$, $c^{g}\in \{0,1\}^{26}$, $s^{r}\in \mathbb{R}^{300}$\\  \hline 
     FC$\rightarrow$1500 for Concat$(z^g,c^g)$, FC$\rightarrow$1500 for $s^r$\\
     FC$\rightarrow$32768, BN, LReLU\\    
     Reshape $16\times16\times 128$\\ \hline 
     $4\times 4$ str.=2  pad.=1 Deconv 64, BN, LReLU\\ 
     $4\times 4$ str.=2  pad.=1 Deconv 1, Tanh \\ \hline
  \end{tabular}
  \end{minipage}
  \quad
  \begin{minipage}[t]{0.45\textwidth}
  \vspace{0pt}
     \begin{tabular}{c}\hline
     {\bf Discriminator} /{\bf Aux. Classifier} \\ \hline
     $x^{r/g} \in \mathbb{R}^{64\times64\times1}$, Expand$(c^{r/g})\in \{0,1\}^{64\times64\times26}$\\   
     $4\times 4$ str.=2  pad.=1 Conv 64, SN, LReLU\\ 
     $4\times 4$ str.=2  pad.=1 Conv 128, SN, LReLU\\ \hline
     FC $\rightarrow$ 1 for Disc.. / FC $\rightarrow$ 1574 for Cl.\\ \hline
     \end{tabular}
 \end{minipage}
 \end{scriptsize}
\end{table}
\section{Experimental Results}\label{sec:results}
\subsection{Font generation specifying a single impression word}

\begin{figure}[!t]
    \centering
   \includegraphics[width=1.0\linewidth]{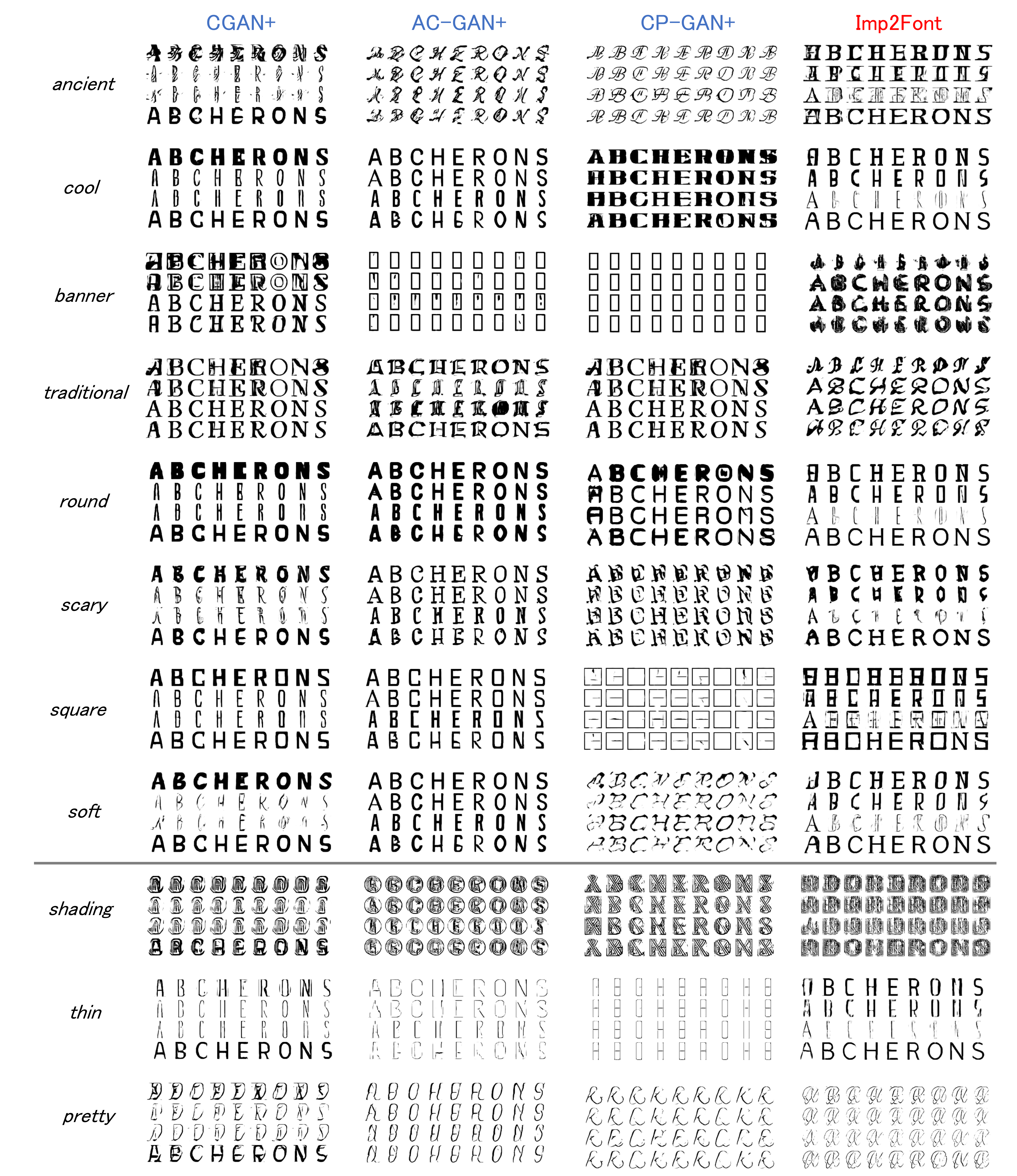}\\
    \caption{Font images generated by specifying a single impression word.}
    \label{fig:generated_fonts_by_single_label}
\end{figure}

Fig.~\ref{fig:generated_fonts_by_single_label} shows the fonts generated by the CGAN+, AC-GAN+, CP-GAN+, and our proposed Imp2Font, conditioned on a single impression word in the $K=1,574$ vocabulary. For each impression, four images are generated for ``ABC'' and ``HERONS''~\footnote{``HERONS'' is a common word to check the font style since it contains sufficient variations of stroke shapes.}, by changing the noise input $z^g$. As noted in Section~\ref{sec:implementation}, the same setup was used for all the models for a fair comparison.
In addition, since each of the competitors can be seen as a subset of Imp2Font, as shown in Fig.~\ref{fig:proposed_model}, their experimental results can be regarded as ablation studies. \par
The result shown in Fig.~\ref{fig:generated_fonts_by_single_label} indicates that Imp2Font could generate font images with the impression specified by the word. For example, all the generated images for {\it square} and {\it round} had more squared and round shapes than others, respectively, while maintaining the diversity of the fonts.
In contrast, the other methods sometimes failed to capture the style specified by the word, as shown in {\it round} and {\it square} for CGAN+.
The result also indicates that images generated by the proposed method maintained the readability of letters in most cases. Meanwhile, the other methods sometimes failed even in the cases where the proposed method succeeds, for example {\it banner} for AC-GAN+ and CP-GAN+ and {\it square} for CP-GAN+. 
Although the proposed method failed in {\it shading} and {\it pretty}, the result of the other methods implies that those words seem to be difficult to reproduce fonts.\par
%
\begin{figure}[t!]
    \centering
   \includegraphics[width=1.0\linewidth]{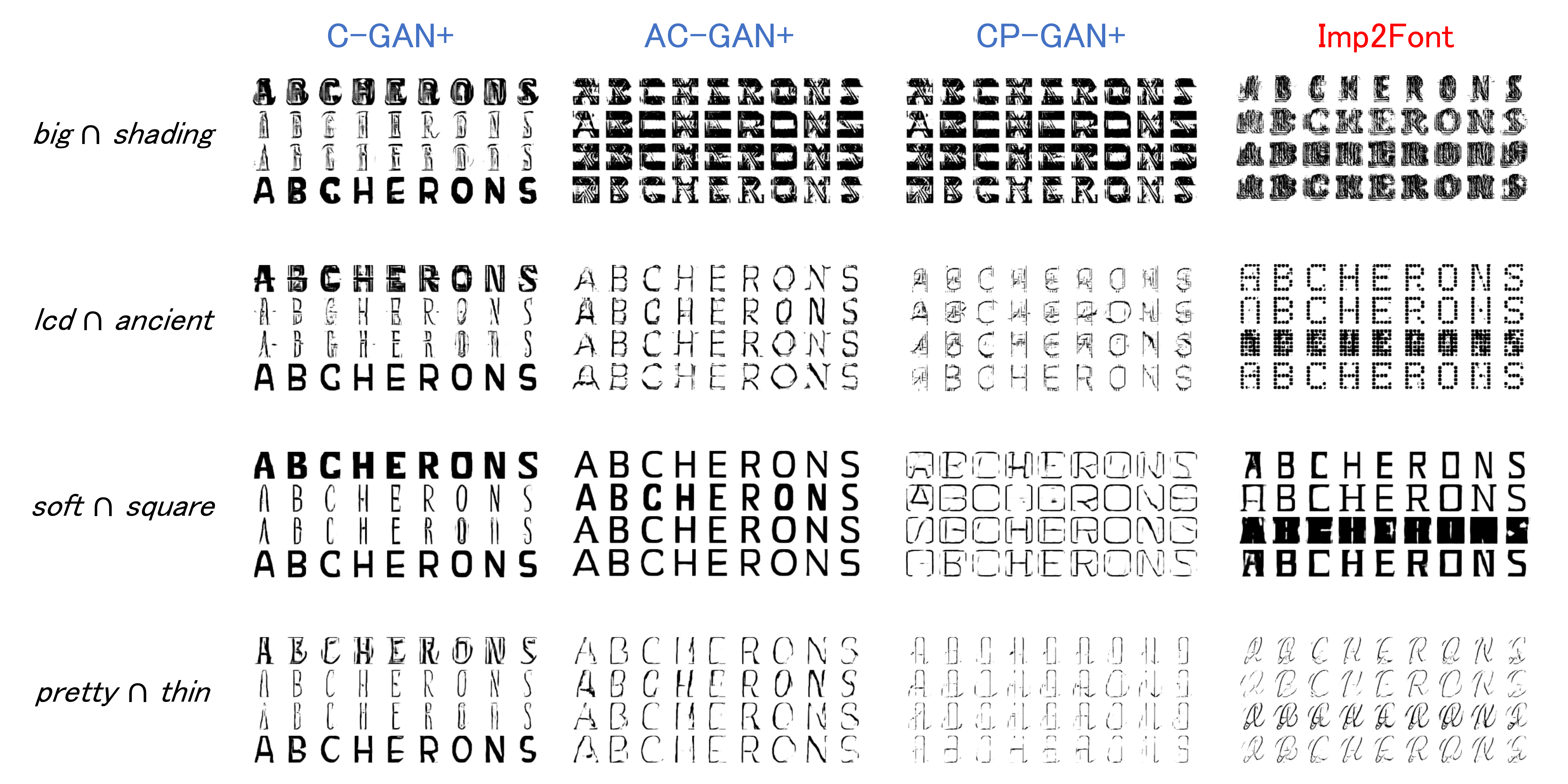}\\[-3mm]
    \caption{Font images generated by specifying multiple impression words.}
    \label{fig:generated_fonts_by_multi_label}
    \bigskip
   \includegraphics[width=1.0\linewidth]{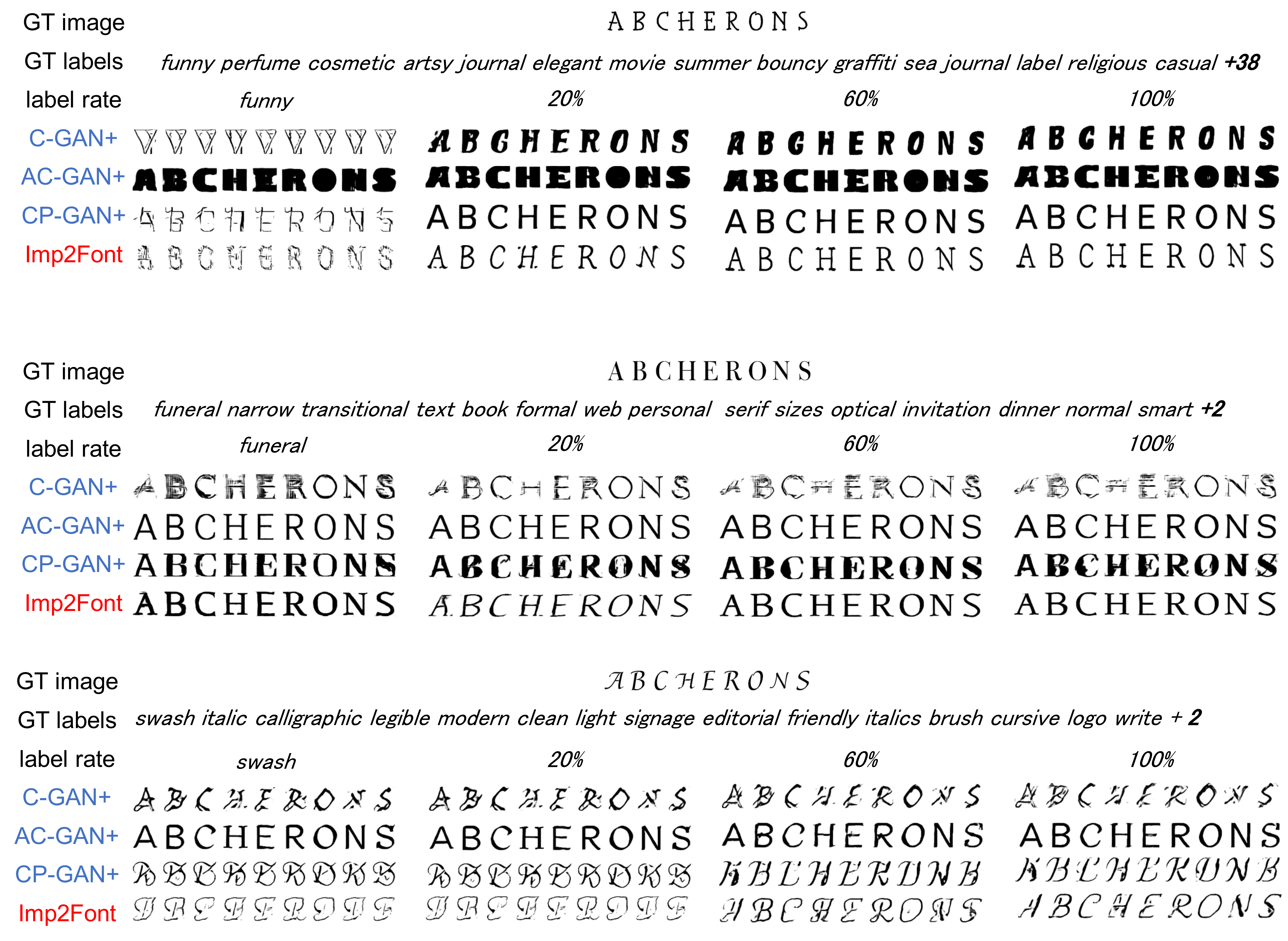}\\[-3mm]
    \caption{Image generated conditional on all impression words in ground truth.}
    \label{fig:comparison_groundtruth}
\end{figure}
\subsection{Font generation specifying multiple impression words}

Fig.~\ref{fig:generated_fonts_by_multi_label}~ shows the font images generated by giving two words as conditions. Our proposed method is very suitable for the generation from multiple impression words; meanwhile the previous methods are not good at this task. This result indicates that Imp2Font could generate font images more visually readable than the ones generated from a single label shown in Fig.~\ref{fig:generated_fonts_by_single_label}. For example, only the generated images by the proposed method for {\it lcd} and {\it ancient} look like dot representations, but they have serif typefaces.
\par

Fig.~\ref{fig:comparison_groundtruth}~ shows the evolution of generated images for increasing numbers of impression words. In those examples, we chose real pairs of an image and the corresponding set of impression words as references and employed each reference set of impression words for generating font images. This result indicates that generated fonts by the proposed method for a reference set of impression words were gradually approaching to the corresponding reference font images; meanwhile the results by the existing methods were far from the reference fonts.\par
\begin{figure}[t]
    \centering
   \includegraphics[width=1.0\linewidth]{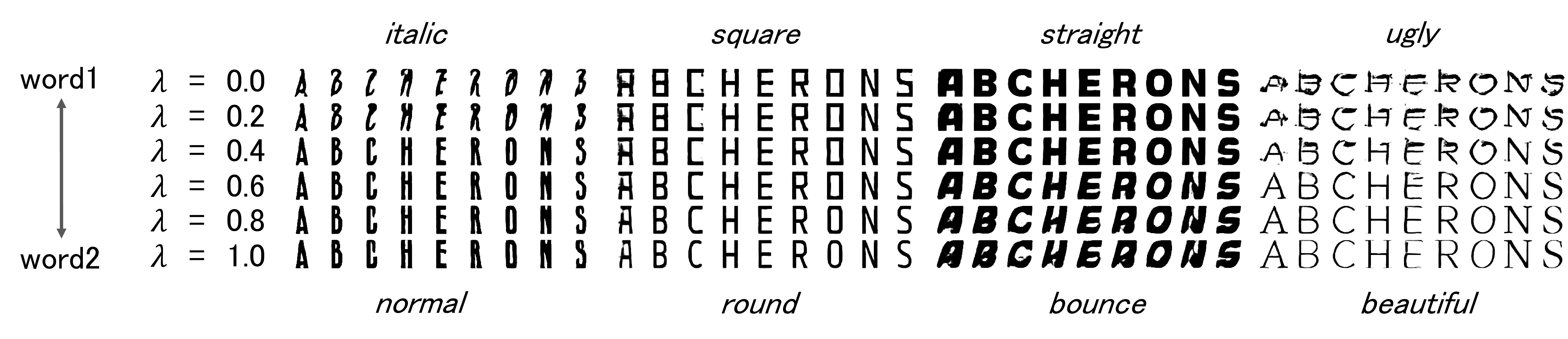}\\[-3mm]
    \caption{Generated glyph images by interpolation between two impression words.}
    \label{fig:interpolation_words}
\end{figure}

Imp2Font can change the strength of impression words as inputs, which means that we can interpolate between different fonts. For example, interpolation $i_{{\it italic}\leftrightarrow{\it normal}}$ between two impression words {\it italic} and {\it normal} can be implemented by $i_{{\it italic}\leftrightarrow{\it normal}}= (1-\lambda)i_{\it italic} + \lambda{i_{\it normal}}$, where $\lambda \in [0,1]$ represents an interpolation coefficient.
Fig.~\ref{fig:interpolation_words}~shows examples of fonts generated with certain interpolation coefficients. In all these examples, the noise $z^g$ is fixed. Imp2Font allows for smooth transitions between two impression words. This result implies that our proposed method could generate a font for a word with multiple impressions, specifying each percentage.\par
\begin{figure}[t]
    \centering
   \includegraphics[width=1.0\linewidth]{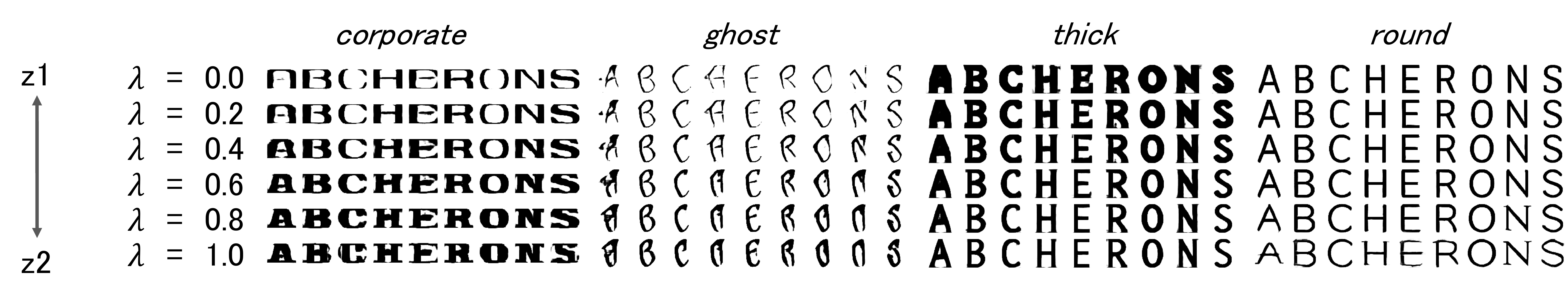}\\[-3mm]
    \caption{Generated glyph images by interpolation between two noise.}
    \label{fig:interpolation_noise}
\end{figure}

Similarly, Imp2Font can control the diversity of fonts even for a fixed set of impression words. We can implement this scheme by interpolating between different noises, i.e., 
$z^g= (1-\lambda)z_1 + \lambda{z_2}.$
Fig.~\ref{fig:interpolation_noise}~ shows examples of fonts generated with certain interpolation coefficients for a fixed single impression word.
It can be seen that the change in noise affected the stroke width and jump in the font style. 

\subsection{Font image generation for unlearned impressions}

\begin{figure}[t]
    \centering
   \includegraphics[width=1.0\linewidth]{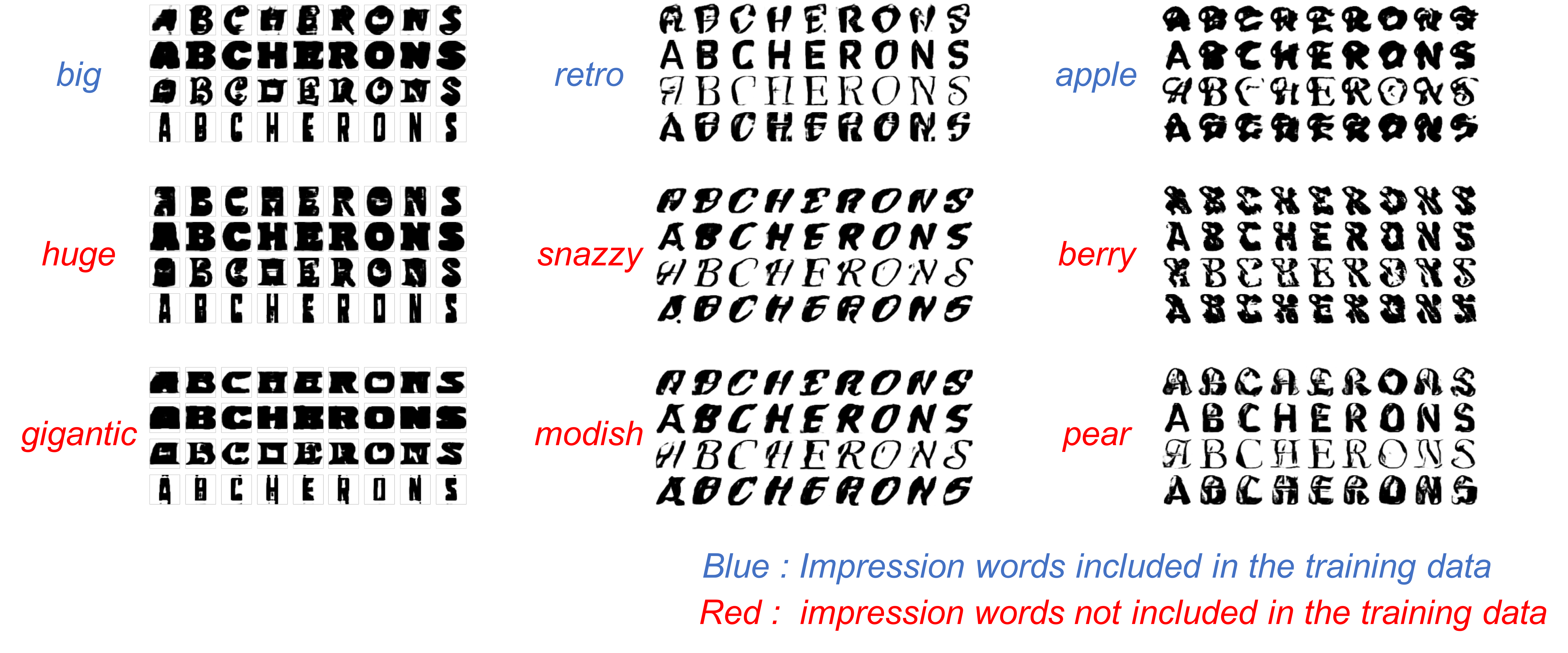}\\[-3mm]
    \caption{Font image generation for unlearned impressions.}
    \label{fig:unlearned-font}
\end{figure}

Imp2Font can also generate font images even for impression words not included in the vocabulary for training. This is because our impression embedding module effectively exploits the knowledge of large-scale external text corpora containing 3 million words through word2vec. This is a unique function of the proposed method that the previous method lacks.
Fig.~\ref{fig:unlearned-font}~ shows the generated font conditioned on the impression words (1) in the vocabulary for learning and (2) those not used for learning but having a similar meaning to the one in the vocabulary for training.
The result shown in this figure indicates that the proposed method could generate fonts even for unlearned impressions, and they were similar to the fonts for similar learned impressions in terms of shapes and styles.



\subsection{Quantitative evaluations}
\begin{table}[!t]
 \caption{Quantitative evaluation results.}
 \label{table:quantitative}
 \centering
  \begin{tabular}{c||r|r|r|r}
    &  CGAN+ & AC-GAN+  & CP-GAN+ & Imp2Font \\ \hline
    $\downarrow$FID & 39.634 & 39.302 & 33.667 & \bf{24.903}\\ 
    $\uparrow$ mAP-train & 1.524 & 1.157 & \bf{1.823} & 1.765\\ 
    $\uparrow$ mAP-test & 1.155 & 1.158 & 1.600 & \bf{1.708}\\ \hline 
  \end{tabular}\\
 
\end{table}

We also compared the proposed method with the previous work in a quantitative manner. We applied two metrics to the evaluation: (1) Frechet inception distance (FID)~\cite{FID} for measuring the distance between distributions of real and generated images. (2) Mean average precision (mAP) that is a standard metric for multi-label prediction. For computing mAPs, we built two additional multi-label ResNet-50 predictors, where one was trained with real images, and their impression words (called {\it pred-real}) and the other was trained with generated images and impression images given for generation (called {\it pred-gen}). In the following, we call mAP for the {\it pred-real} model and generated test images {\it mAP-train}, and the one for the {\it pred-gen} and real test images {\it mAP-test}, respectively.

Table~\ref{table:quantitative}~shows the quantitative results.
The results indicate that the proposed method greatly outperformed the others for almost all the cases. More specifically, the proposed method and CP-GAN+ improved mAPs against CGAN+ and AC-GAN+. This implies that the introduction of the impression embedding module that provides soft constraints for generators was effective for our task. Also, we can see that the proposed method greatly improved FID, which implies that the proposed method can generate high-quality font images while maintaining the diversity of fonts. Note that the mAP-train of Imp2Font is slightly lower than CP-GAN+; one possible reason is that the word embedding performance is still not perfect. Currently, Imp2Font employs word2vec, which is based on the distributional hypothesis and thus has a limitation that words with very different meanings (even antonyms) have a risk of similar embedding results. Using advanced embedding methods will easily enhance the performance of Imp2Font.


\section{Conclusion}\label{sec:conclusion}
This paper proposes Impressions2Font (Imp2Font), a novel conditional GAN, and enables generating font images with specific impressions. Imp2Font accepts an arbitrary number of impression words as its condition to generate the font. Internally, by an impression embedding module, the impression words are converted to a set of semantic vectors by a word embedding method and then unified into a single vector using likelihood values of individual impressions as weights. This single vector is used as a soft condition for the generator. Qualitative and quantitative evaluations prove the high quality of the generated images. Especially, it is proved that giving more impression words will help to generate the expected font shape accurately. Moreover, it is also proved that Imp2Font can accept even {\em unlearned} impression words by using the flexibility of the impression embedding module. \par
Future work will focus on incorporating different word embedding techniques instead of word2vec. Especially if we can realize a new word embedding technique grounded by incorporating the shape-semantic relationship between fonts and their impressions, like sound-word2vec~\cite{sw2v} and color word2vec~\cite{cw2v}, it is more appropriate for our framework. 
We also plan to conduct evaluation experiments for understanding not only the legibility but also the impression of the generated font images.
\section*{Acknowledgment}
This work was supported by JSPS KAKENHI Grant Number JP17H06100.
%
%
%
\bibliographystyle{splncs04}
\bibliography{icdar}
%





\end{document}